\documentclass{article} % For LaTeX2e
\usepackage[final]{colm2025_conference}

\usepackage{microtype}
\usepackage{hyperref}
\usepackage{url}
\usepackage{booktabs}
\usepackage{graphicx}
\usepackage{enumitem}
\usepackage{float}
\usepackage[most]{tcolorbox} 
\usepackage{lineno}

\definecolor{darkblue}{rgb}{0, 0, 0.5}
\hypersetup{colorlinks=true, citecolor=darkblue, linkcolor=darkblue, urlcolor=darkblue}

\title{Sarc7: Evaluating Sarcasm Detection and Generation with Seven Types and Emotion-Informed Techniques}

\author{
\textbf{Raina Gao}\thanks{Equal contribution.} \quad
\textbf{Alyssa Jeong}\footnotemark[1] \quad
\textbf{Lang Xiong} \quad
\textbf{Yicheng Fu} \\
\textbf{Sean O'Brien} \quad
\textbf{Vasu Sharma} \quad
\textbf{Kevin Zhu} \\
Algoverse AI Research \\
\texttt{sean@algoverse.us, kevin@algoverse.us} \\
}
\begin{document}

\maketitle

\begin{abstract}
Sarcasm is a complex linguistic and pragmatic phenomenon where expressions convey meanings that contrast with their literal interpretations, requiring sensitivity to the speaker's intent and context. Misinterpreting sarcasm in collaborative human–AI settings can lead to under- or overreliance on LLM outputs, with consequences ranging from breakdowns in communication to critical safety failures. We introduce \textbf{Sarc7}, a benchmark for fine-grained sarcasm evaluation based on the MUStARD dataset, annotated with seven pragmatically defined sarcasm types: self-deprecating, brooding, deadpan, polite, obnoxious, raging, and manic. These categories are adapted from prior linguistic work and used to create a structured dataset suitable for LLM evaluation. For classification, we evaluate multiple prompting strategies—zero-shot, few-shot, chain-of-thought (CoT), and a novel emotion-based technique—across five major LLMs. Emotion-based prompting yields the highest macro-averaged F1 score of 0.3664 (Gemini 2.5), outperforming CoT for several models and demonstrating its effectiveness in sarcasm type recognition. For sarcasm generation, we design structured prompts using fixed values across four sarcasm-relevant dimensions: incongruity, shock value, context dependency, and emotion. Using Claude 3.5 Sonnet, this approach produces more subtype-aligned outputs, with human evaluators preferring emotion-based generations 38.46\% more often than zero-shot baselines. Sarc7 offers a foundation for evaluating nuanced sarcasm understanding and controllable generation in LLMs, pushing beyond binary classification toward interpretable, emotion-informed language modeling.

\end{abstract}

\section{Introduction}
Sarcasm is defined as the use of remarks that convey the opposite of their literal meaning. Understanding sarcasm requires an intuitive grasp of humor and social cues, posing a challenge for natural language processing (NLP) tasks such as human-like conversation \citep{yao2024sarcasm, gole2024sarcasm}. Sarcasm is a pragmatic act, where meaning depends not only on words but also on speaker intent, emotional tone, and shared context. Large language models (LLMs) generally perform poorly on sarcasm classification and generation tasks due to the subtlety and context dependence of sarcastic language \cite{yao2024sarcasm}. Traditional sentiment analysis and machine learning techniques also struggle with these challenges. 
This work introduces a novel sarcasm benchmark grounded in the seven recognized types of sarcasm and proposes an emotion-based approach for both classification and generation. We examine whether LLMs can demonstrate pragmatic reasoning. In contrast to prior rule-based and template-driven methods, which often produced rigid outputs \cite{zhang2024sarcasmbench}, and even more recent deep learning models that still fall short in capturing subtlety and social nuance \cite{gole2024sarcasm}, our technique aims to improve contextual relevance and expressive range in sarcastic generation. 
Sarcasm often serves as a subtle guardrail in human dialogue: 
misreading it can cause an AI “collaborator” either to underreact 
(e.g. miss a warning hidden in irony) or to overreact (e.g. execute 
an unintended action on a sarcastic command). In high-stakes 
settings—negotiations, medical advice, compliance checks—such 
misinterpretations can undermine both safety and user trust.  
Sarc7 thus not only measures pragmatic competence but also 
lays the groundwork for accountability in human–LLM teams, 
helping to modulate reliance on AI by flagging when ironic intent 
makes a literal interpretation hazardous.  

\section{Related Work}

Previously, SarcasmBench \cite{zhang2024sarcasmbench} established benchmarks for binary sarcasm classification by evaluating state-of-the-art (SOTA) large language models (LLMs) and pretrained language models (PLMs). \cite{leggitt2000emotional,biswas2019computational}. Lamb (2011) first introduced a seven‐type classification of sarcasm based on observational studies of classroom discourse. \cite{qasimcritical} then refined these categories into operational definitions tailored for social‐interview data, providing clear examples and criteria. \cite{zuhri2022irony} subsequently applied this refined taxonomy in an irony and sarcasm detection system for public‐figure speech. Sarc7 translates those categories into concrete definitions and detailed annotation guidelines to construct and evaluate our Sarc7 benchmark for LLMs.

\textbf{Reliance and Accountability.}  
Studies show that users often over-rely on LLM outputs, even when they are unreliable \cite{lee2023overtrust,}.  
Conversely, under-reliance prevents users from leveraging LLM efficiency \cite{wang2022human}.  
Mechanisms such as local explanations or confidence warnings can help calibrate trust \cite{zhang2024explainability}.  
Sarc7’s fine-grained labels and stepwise CoT explanations directly support these mitigation strategies by:  
(1) highlighting when sarcastic intent may mislead an agent, and  
(2) providing interpretable rationales to apportion accountability for downstream decisions.

\textbf{Sarcasm Classification:}
\cite{riloff2013sarcasm} introduced a sentiment-contrast framework for binary sarcasm detection, flagging instances where positive wording clashes with negatively described contexts. Recent advances have focused on structured prompting techniques that use pragmatic reasoning to enhance sarcasm detection \cite{lee2024pragmatic}. Approaches such as pragmatic metacognitive prompting method (PMP) have improved model performance by making sarcasm inference more explicit \cite{yao2024sarcasm, lee2024pragmatic}. Furthermore, recent studies have shown that integrating commonsense, knowledge, and attention mechanisms help models identify subtleties in sarcastic statements \cite{zhuang2025multi}. These methods show that guiding LLMs with structured signals can help them better understand the nuances of sarcastic statements. 

\textbf{Sarcasm Generation:}
Recent studies have introduced controlled generation methods to guide LLMs toward producing sarcastic statements using contradiction strategies and dialogue cues \cite{zhang2024sarcasmbench, helal2024contextual}. Structured prompting and contradiction-based strategies have shown to improve sarcasm generation. Some methods guide LLMs by introducing contrast between expected and actual meanings or using contextual dialogue cues for coherence \cite{zhang2024sarcasmbench, helal2024contextual, skalicky2018linguistic}. However, existing techniques struggle with controlling sarcasm levels and aligning them with contextual incongruence, shock value, and prior context dependency.
\section{Methods}

\begin{figure}[h]
  \centering
  \includegraphics[width=0.45\columnwidth]{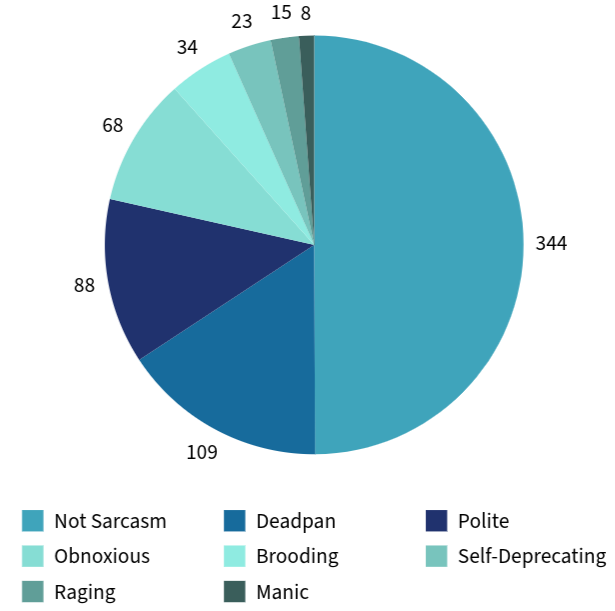}
  \caption{Distribution of Annotation Labels in the Dataset.}
  \label{fig:piechart}
\end{figure}
\subsection{Benchmark Construction}

We introduce \textbf{Sarc7}, a novel benchmark for fine-grained sarcasm classification and generation. Building on the MUStARD dataset \citep{castro-etal-2019-towards}, which provides binary sarcasm annotations for short dialogue segments, we manually annotated each sarcastic utterance with one of seven distinct sarcasm types: \textit{self-deprecating}, \textit{brooding}, \textit{deadpan}, \textit{polite}, \textit{obnoxious}, \textit{raging}, and \textit{manic}.

These seven categories are inspired by the linguistic taxonomy proposed in \citet{qasimcritical}, which identified common sarcasm types based on pragmatic and affective features. Our contribution lies in implementing these types of sarcasm for computational annotation. We defined each type using precise, example-grounded criteria suitable for large language model evaluation, and we applied this schema to build the first sarcasm benchmark that captures this level of granularity.

\begin{figure*}[h]
  \includegraphics[width=\textwidth,height=4cm]{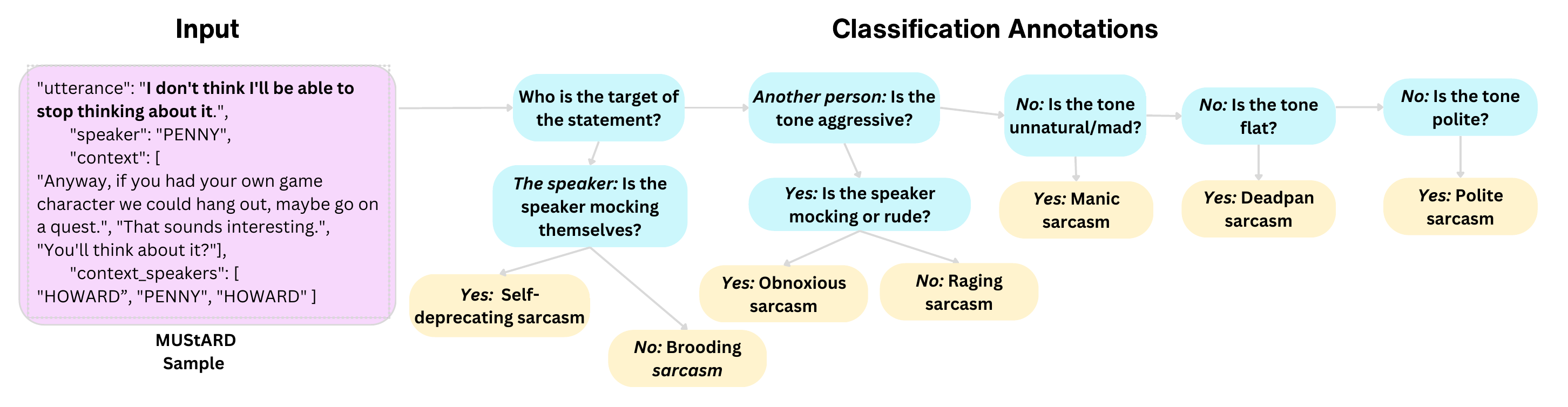}
    \caption{Flowchart of the Step-by-Step Process for Sarcasm Classification Annotation}
  \label{fig:flowchart}
\end{figure*}

\subsection{Annotation Methodology}
Each sarcastic utterance in the MUStARD dataset (n=690) was independently labeled by four trained annotators using the seven sarcasm subtypes defined in Sarc7. Annotators were instructed to consider pragmatic cues and received detailed definitions and examples of each category (see Table~\ref{tab:sarcasm_types}) to ensure consistent interpretation. The annotation process is illustrated in Figure~\ref{fig:flowchart}.
\begin{itemize}
    \item Each utterance was first labeled independently by all four annotators.
    \item If at least three annotators agreed on the same label, that label was accepted as the final annotation.
    \item In cases with no 3-out-of-4 agreement, a consensus discussion was held between annotators, with a final decision made by majority vote.
\end{itemize}

To quantify the reliability of our 3-of-4 consensus labels, we recruited a fifth trained annotator to re-label all utterances independently. We then computed Cohen’s kappa between the majority vote (from the original four annotators) and this fifth annotator’s labels. The resulting Cohen’s $\kappa$ = 0.6694 indicates substantial agreement according to \cite{landis1977measurement} scale. The macro-averaged precision, recall, and F1 for this human comparison were 0.6586, 0.6847, and 0.6663, respectively.  This provides further evidence that our annotation schema is both consistent and replicable. Even for trained readers, \textbf{brooding}, \textbf{deadpan}, and \textbf{polite} sarcasm proved the most challenging to label consistently, establishing realistic upper bounds for model performance on these subtypes.

Figure \ref{fig:piechart} shows the distribution of the seven annotated sarcasm types. The resulting Sarc7 benchmark supports two tasks: (1) multi-class sarcasm classification, and (2) sarcasm-type-conditioned generation. These tasks allow for more fine-grained evaluation of sarcasm understanding in large language models.

\subsection{Task Definition}

We define two primary evaluation tasks:

\begin{itemize}[noitemsep, topsep=0pt, parsep=0pt, partopsep=0pt]
    \item \textbf{Sarcasm Classification}: Given a sarcastic utterance and its dialogue context, correctly predict the dominant sarcasm type from among the seven annotated categories.
    \item \textbf{Sarcasm Generation}: Generate a sarcastic utterance consistent with one of the 7 types of sarcasm. Table~\ref{tab:sarcasm_types} outlines definitions for each sarcasm category in the Sarc7 benchmark.
\end{itemize}

\begin{table}[h]
\small
\centering
\begin{tabular}{p{2.4cm} p{4.2cm} p{5.3cm}}
\toprule
\textbf{Type} & \textbf{Definition} & \textbf{Example} \\
\midrule
Self-deprecating & Mocking oneself in a humorous or critical way. & “Oh yeah, I’m a genius — I only failed twice!” \\
Brooding & Passive-aggressive frustration masked by politeness. & “Sure, I’d love to stay late again — who needs weekends?” \\
Deadpan & Sarcasm delivered in a flat, emotionless tone. & “That’s just the best news I’ve heard all day.” \\
Polite & Insincere compliments or overly courteous remarks. & “Wow, what an \textit{interesting} outfit you’ve chosen.” \\
Obnoxious & Rude or provocative sarcasm aimed at others. & “Nice driving! Did you get your license in a cereal box?” \\
Raging & Intense, exaggerated sarcasm expressing anger. & “Of course! I \textit{love} being yelled at in meetings!” \\
Manic & Overenthusiastic, erratic sarcasm with chaotic tone. & “This is AMAZING! Who needs food or sleep anyway?!” \\
\bottomrule
\end{tabular}
\caption{Operational Definitions and Examples of the Seven Sarcasm Types used in Sarc7}
\label{tab:sarcasm_types}
\end{table}

\subsection{Baseline Classification}
Our baseline testing focused on zero-shot, few-shot, and CoT prompting. For generations, baseline outputs were produced using a zero-shot prompt, without structured control over dimensions. These baselines were evaluated by a human grader based on accuracy of sarcasm type and emotion.

\subsection{Emotion-Based Prompting}
Our emotion‐based prompting goes beyond traditional sentiment analysis by leveraging discrete emotion categories rather than coarse positive/negative polarity. This method captures pragmatic incongruity through emotional mismatches, approximating listener inference. Whereas sentiment classifiers typically flag a mismatch between overall sentiment and context \cite{riloff2013sarcasm}, our approach leverages the six basic emotions identified by American psychologist Paul Ekman: happiness, sadness, anger, fear, disgust, and surprise \cite{ekman1992there}. Our emotion-based prompting technique consists of three main steps: 1) Categorize the emotion of the context. 2) Classify the emotion of the utterance. 3) Identify the sarcasm based on the incongruity of the emotional situation. By comparing these two emotion labels, we capture nuanced contrasts—such as polite sarcasm pairing happiness with a neutral situation or obnoxious sarcasm pairing neutral context with a superficially disgusting utterance—that a simple positive/negative split cannot distinguish. This fine‐grained emotional reasoning provides a clear advantage for multi‐class sarcasm classification: it supplies subtype‐specific cues (e.g., “raging” sarcasm requires anger, “manic” requires surprise or happiness) and thus helps disambiguate among several closely related sarcasm types rather than collapsing them all into a single sarcastic category. By explicitly unpacking the speaker’s emotional state before assigning a sarcasm label, emotion-based prompts reduce the model’s bias toward the dominant “not sarcasm” class and guard against failure modes where agents take ironic instructions literally. This local explanation step also allows downstream systems to log and audit each pragmatic cue—enabling accountability if misclassification leads to unsafe decisions.  

%To assess the accuracy of LLMs in comparison to human understanding of sarcasm, we conducted a human classification task. One annotator was tasked with classifying all data points in the MUStARD dataset after they had been initially annotated with sarcasm types. 
%\subsection{Model Selection}

\begin{table}[h]
\centering
\small
\begin{tabular}{lcccc}
\toprule
\textbf{Subtype} & \textbf{Incongruity (1–10)} & \textbf{Shock Value} & \textbf{Context Dependency} & \textbf{Emotion} \\
\midrule
Self‐deprecating       & 3–5 & low    & medium & sadness    \\
Brooding               & 5–7 & medium & medium & anger      \\
Deadpan                & 4–6 & low    & high   & neutral    \\
Polite                 & 3–5 & low    & medium & happiness  \\
Obnoxious              & 6–9 & high   & low    & disgust    \\
Raging                 & 7–9 & high   & low    & anger      \\
Manic                  & 5–7 & high   & medium & surprise   \\
\bottomrule
\end{tabular}
\caption{Dimension Settings and Target Emotion for Each Sarcasm Subtype used in our Emotion‐based Prompting.}
\label{tab:dim-settings}
\end{table}
\subsection{Generation Dimensions}

Our approach moves beyond general sarcasm generation by conditioning the model on four controllable pragmatic dimensions intended to guide the tone, intensity, and context of the output:

\begin{itemize}[noitemsep, topsep=0pt, parsep=0pt, partopsep=0pt]
    \item \textbf{Incongruity}: Degree of semantic mismatch (1-10).
    \item \textbf{Shock Value}: Intensity of sarcasm.
    \item \textbf{Context Dependency}: Reliance on conversational history.
    \item \textbf{Emotion}: One of Ekman’s six basic emotions (e.g., anger, sadness).
\end{itemize}

\begin{figure}[h]
  \centering
  \includegraphics[width=0.8\columnwidth]{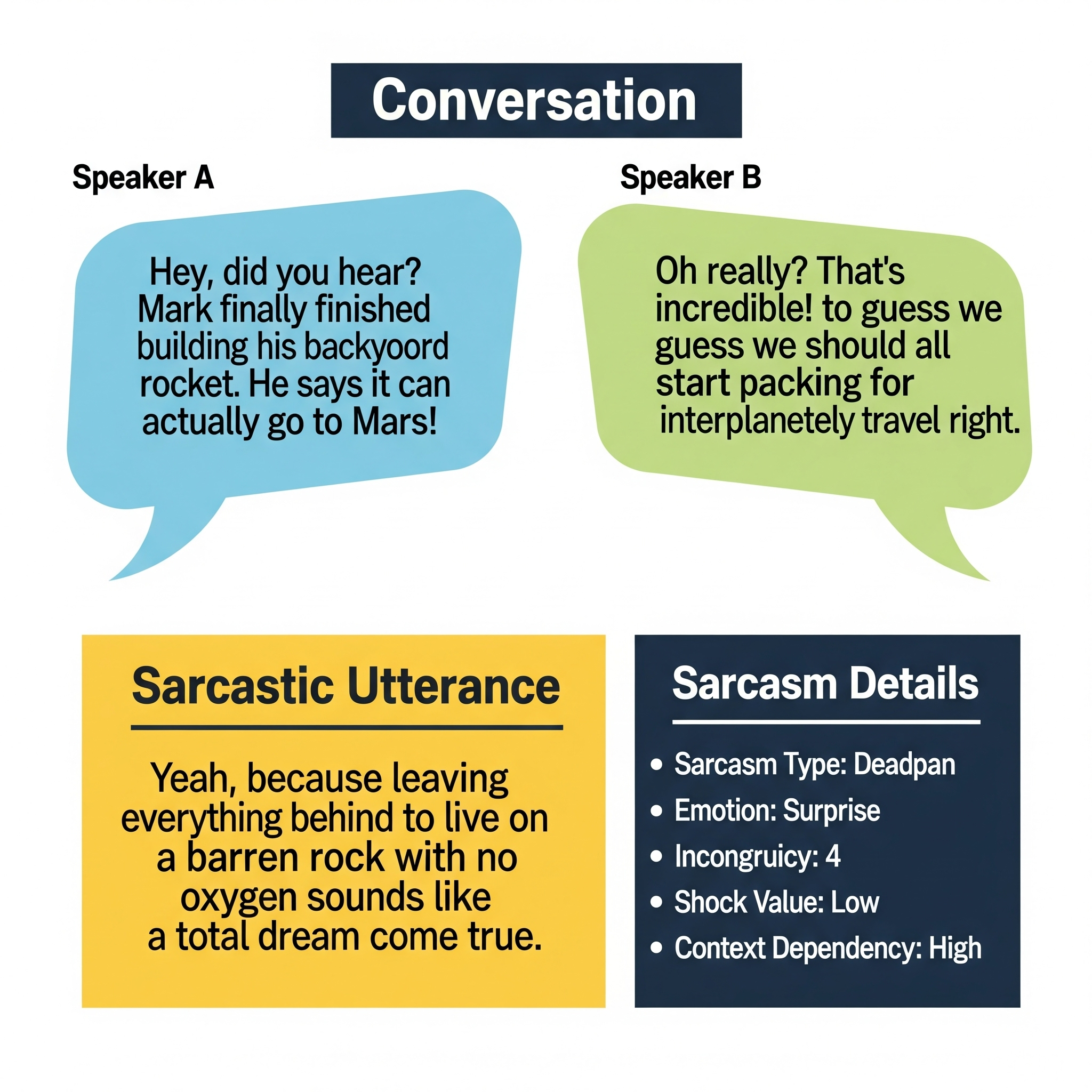}
    \caption{Sample Output Using Emotion-based Generation Method}
  \label{fig:experiments}
\end{figure}

Rather than tuning these dimensions dynamically, we assigned fixed values for each subtype based on our intuitive understanding (see Table~\ref{tab:dim-settings}). By anchoring each generation to these abstract but interpretable cues, we observed improved alignment between the generated outputs and their intended sarcasm type. This structured prompting approach helps control for variation in tone and emotional affect, resulting in more consistent and subtype-specific sarcasm generation. A sample output from this technique is shown in Figure~\ref{fig:experiments}.

\section{Experiments}

\subsection{Model Selection}
We evaluate several state-of-the-art language models on our proposed sarcasm benchmark, including GPT-4o \cite{openai2024gpt4o}, Claude 3.5 Sonnet \cite{anthropic2024claude3}, Gemini 2.5 \cite{deepmind2023gemini}, Qwen 2.5 \cite{qwen2024technical}, and Llama 4 Maverick \cite{llama4maverick}. 

% These models represent a diverse set of architectures and training paradigms, enabling a broad assessment of sarcasm classification and generation capabilities. 

\subsection{Evaluation}
We evaluated classification by comparing model predictions to human-annotated labels across seven sarcasm types. For generation, Claude 3.5 Sonnet produced 100 sarcastic statements per prompting method, each rated by a human for sarcasm type accuracy.

\section{Results and Discussion}
\begin{table*}[h]
  \centering
  \begin{tabular}{lcccc}
    \toprule
    \textbf{Model} & \textbf{0-shot} & \textbf{Few-shot} & \textbf{CoT} & \textbf{Emotion-based} \\
    \midrule
    GPT-4o             & 47.73\% & 50.29\% & \textbf{55.07\%} & 48.94\% \\
    Claude 3.5 Sonnet  & 51.16\% & 52.61\% &  \textbf{57.10\%} & 52.32\% \\
    Qwen 2.5           & 41.45\% & \textbf{46.96\%} & 46.09\% & 45.94\% \\
    Llama-4 Maverick   & 34.20\% & 35.51\% & \textbf{50.29\%} & 49.86\% \\
    Gemini 2.5         & 46.81\% & 47.97\% & \textbf{53.04\%} & 52.03\% \\
    \bottomrule
  \end{tabular}
  \caption{Classification Accuracy Across Models and Prompting Techniques}
  \label{tab:classification-results}
\end{table*}
\begin{table*}[h]
  \centering
  \begin{tabular}{lcccc}
    \toprule
    \textbf{Model} & \textbf{0-shot F1} & \textbf{Few-shot F1} & \textbf{CoT F1} & \textbf{Emotion-based F1} \\
    \midrule
    GPT-4o             & 0.2089 & \textbf{0.3255} & 0.2674 & 0.2233 \\
    Claude 3.5 Sonnet  & 0.2964 & 0.3487 & 0.2471 & \textbf{0.3487} \\
    Qwen 2.5           & 0.2116 & 0.2075 & 0.2052 & \textbf{0.2124} \\
    Llama-4 Maverick   & 0.2184 & 0.2340 & 0.2040 & \textbf{0.2841} \\
    Gemini 2.5         & 0.2760 & 0.3274 & 0.3141 & \textbf{0.3664} \\
    \bottomrule
  \end{tabular}
  \caption{Macro-averaged F1 scores of Models Across Prompting Techniques.}
  \label{tab:f1-results}
\end{table*}
\subsection{Classification Results}
Across all evaluated prompting techniques, CoT prompting consistently outperformed zero-shot, few-shot, and emotion-based approaches in sarcasm classification. Table \ref{tab:classification-results} shows its superior results compared to other methods in almost every model. In terms of macro-averaged F1 score, emotion-based prompting outperformed zero-shot, few-shot, and CoT prompting. As shown in Table \ref{tab:f1-results}, Gemini 2.5 achieved the highest F1 score overall under emotion-based prompting, with Claude 3.5 Sonnet, Llama-4 Maverick, and Qwen 2.5 also seeing gains relative to their CoT performance. While CoT prompting remains strong in absolute accuracy and reasoning through ambiguous cases, emotion-based prompting demonstrated greater ability to generalize across sarcasm types, especially those associated with emotional signals. This improvement is particularly important given the dataset’s class imbalance. Since types like “Deadpan” appear more frequently than others such as “Manic” or “Polite,” raw accuracy metrics may disproportionately reflect dominant class performance. Macro-averaged F1 provides a more balanced evaluation by weighting each class equally. The higher F1 scores observed under emotion-based prompting suggest that emotional cues may help LLMs better distinguish between low-frequency categories, even in the absence of detailed reasoning steps.

\begin{figure}[h]
    \centering
  \includegraphics[width=0.6\columnwidth]{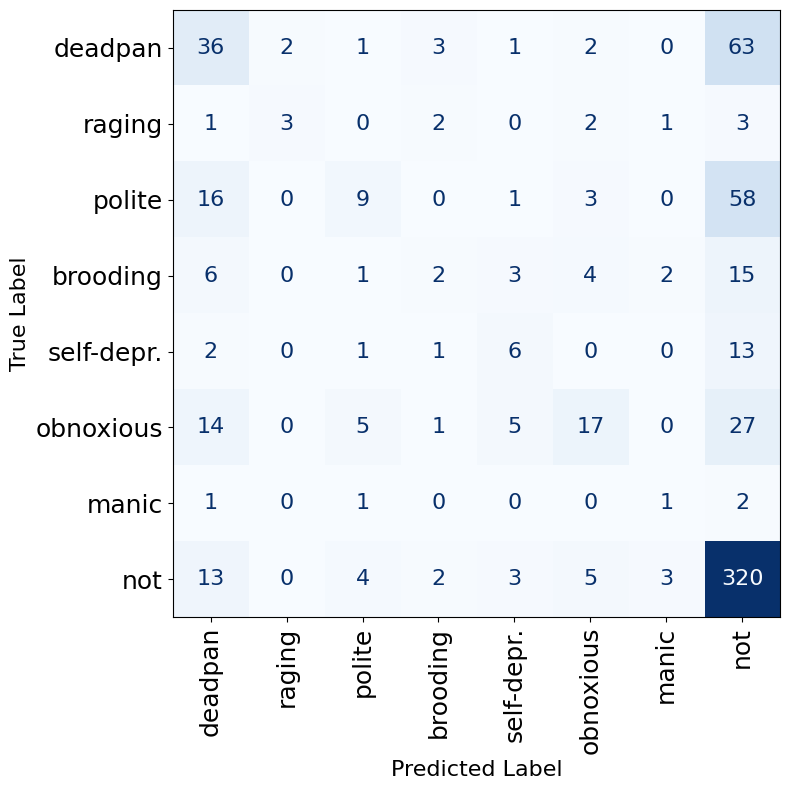}
    \caption{Confusion Matrix for Claude 3.5 Sonnet using CoT.}
  \label{fig:confusion}
\end{figure}

\subsection{Classification Confusion Analysis}

While models showed moderate success identifying sarcastic utterances, they struggled to accurately categorize specific sarcasm types. Figure \ref{fig:confusion} shows that most models, including GPT4o, Claude 3.5 Sonnet, and Gemini 2.5, frequently defaulted to labeling content as either "not sarcastic" or "deadpan sarcasm" when uncertain. Deadpan emerged as the most frequent misclassification across all sarcasm types, underscoring its role as a default or fallback label in ambiguous cases. This trend reveals a key limitation: although LLMs can sometimes detect cues associated with sarcastic tone, they often conflate subtle, flat, or ambiguous language with sarcasm—even when none is present. The frequent misclassification of non-sarcastic utterances as "deadpan" indicates that models are over-reliant on surface-level features such as flat affect or contrastive phrasing, rather than grounded pragmatic reasoning. As a result, fine-grained differentiation among sarcasm subtypes remains a substantial challenge. Improving model sensitivity to context and disambiguation of neutral tone from intentional sarcasm is critical for more accurate multi-class sarcasm detection.

\begin{table}[h]
  \centering
  \small
  \begin{tabular}{lccc}
    \toprule
    \textbf{Subtype}            & \textbf{CoT} & \textbf{Emotion-based} & \textbf{Human} \\
    \midrule
    Brooding sarcasm           &  6.06\%      &  9.09\%                & 39.39\%  \\
    Deadpan sarcasm            & 33.03\%      & 50.46\%                & 55.45\%  \\
    Polite sarcasm             & 10.34\%      & 33.33\%                & 57.30\%  \\
    Manic sarcasm              & 20.00\%      & 20.00\%                & 75.00\%  \\
    Obnoxious sarcasm          & 24.64\%      & 39.13\%                & 67.14\%  \\
    Raging sarcasm             & 25.00\%      & 41.67\%                & 71.43\%  \\
    Self‐deprecating sarcasm   & 26.09\%      & 34.78\%                & 86.96\%  \\
    Not sarcasm                & 91.17\%      & 66.38\%                & 95.04\%  \\
    \bottomrule
  \end{tabular}
  \caption{Per‐class Accuracy for Claude 3.5 using CoT vs.\ Emotion-based Prompting, Alongside Human Agreement.}
  \label{tab:class-acc-comparison}
\end{table}

Table~\ref{tab:class-acc-comparison} shows that emotion-based prompting yields consistent relative improvements over CoT prompting, though absolute accuracy remains below the human ceiling.  In particular, brooding gains +3.04\%, polite +23.0 \%, deadpan +17.47 \%, and raging +16.67 \%, demonstrating that emotion cues help disambiguate more subtle tones.  Conversely, “not sarcasm” drops by –24.82 \%, indicating that adding emotion information can sometimes introduce noise for clear non‐sarcastic cases.  These shifts confirm that emotion‐based prompts move the model closer to human‐level nuance on mid‐difficulty classes, but the largest remaining gaps still align with the hardest human distinctions—especially brooding, deadpan, and polite sarcasm—suggesting the need for richer contextual and pragmatic reasoning beyond fixed emotion settings. From a pragmatic standpoint, these patterns show that fixed emotion cues can help LLMs avoid the default “deadpan” trap in nuanced cases, but true conversational implicature often depends on richer context and iterative hypothesis testing. The persistent gaps on brooding, deadpan, and polite highlight subtypes whose disambiguation relies heavily on prosodic and interpersonal cues—elements our current text-only prompting cannot capture. 

\subsection{Prompt Technique Analysis}

Emotion‐based prompting, which explicitly models the listener’s pragmatic hypothesis—“What emotion is intended here?”—yields higher macro‐F1, demonstrating better performance on low‐frequency sarcasm subtypes, indicating that discrete emotional cues guide LLMs toward the correct implicature when literal context is sparse. In contrast, CoT prompting excels at overall accuracy by simulating pragmatic inference, but can overlook subtler emotional distinctions; this trade‐off underscores the need to balance structured reasoning with direct emotion signals when modeling conversational implicature in multi‐class sarcasm. 
From a planning perspective, our CoT prompts serve as local explanations: each step (tone→mismatch→type) lays bare the model’s hypothesis testing, similar to feature‐level attributions in vision. Emotion‐based prompts deliver a more compact textual rationale—choosing context‐emotion and utterance‐emotion—that surfaces the primary pragmatic signal. These complementary explanation styles enhance 
both model effectiveness (by guiding correct inference on low‐frequency subtypes) and safety (by making each reasoning path auditable).

\subsection{Qualitative Error Analysis}

Despite strong binary performance, models often misclassify playful language as sarcasm. Consider the following example:

\begin{quote}
\textbf{Utterance:} \texttt{A lane frequented by liars. Like you, you big liar!} \\
\textbf{Context:} \texttt{HOWARD: I just Googled "foo-foo little dogs." \\
HOWARD: (Skype ringing) It's Raj. Stay quiet. \\
HOWARD: (chuckles): Hey! \\
Bad timing. \\
Bernadette just took Cinnamon out for a walk. \\
RAJ: Hmm. Interesting. \\
Did they take a walk down Liars' Lane? \\
HOWARD: What?}
\end{quote}

The true label is \textit{not sarcastic}, yet all models predicted \textit{obnoxious sarcasm}. The CoT prompt overemphasized surface-level markers such as exaggeration and contradiction, failing to consider the light tone of the exchange. Similarly, the emotion-based prompt misclassified the utterance by identifying "disgust" due to literal wording, despite the playful social context. These errors highlight a broader limitation: while structured prompting improves reasoning, both CoT and emotion-based methods lack sensitivity to pragmatic cues and interpersonal intent in conversational sarcasm.

\begin{table}[H]
\centering
\begin{tabular}{l c}
  \toprule
  \textbf{Prompt}        & \textbf{Successful Generation} \\
  \midrule
  Zero-shot              & 52/100 \\
  \textbf{Emotion-based} & \textbf{72/100} \\
  \bottomrule
\end{tabular}
\caption{Generation Evaluation Scores}
\label{tab:generation-eval}
\end{table}

\subsection{Generation Results and Analysis}

Emotion-based prompting generated more accurate sarcasm types. Table \ref{tab:generation-eval} shows a 38.42\% increase in accuracy using the emotion-based structure compared to the baseline model.

For example, when prompted for raging sarcasm zero-shot prompting produced a neutral response:

\textit{"Oh, absolutely! I only stayed up until 3 AM because sleep is just so overrated, right?"}

The emotion-based prompt with angry context and high shock value generated:

\textit{"Isn't that just fantastic? I mean, who wouldn't want to spend an entire day writing reports on how well we walk from our desks to the restroom? It's a dream come true!"}

The baseline prompt's neutral context made it difficult to generate raging sarcasm, likely confusing it with deadpan due to the absence of anger cues. However, our emotion-based prompt was able to identify the anger in the statement and appropriately express it in its response. By structuring generation through pragmatic dimensions like context dependency and incongruity, our method implicitly guides the model to replicate speaker goals. See Appendix \ref{appendix:b} for examples' context. Notably, brooding and manic sarcasm were the toughest for LLMs to generate. Brooding depends on a courteous veneer masking genuine frustration, a nuance carried by tone and pacing, not keywords, so single-turn prompts slip into blunt reproach. Manic sarcasm requires sustained, erratic enthusiasm that signals insincerity through vocal intensity; without prosody, models fall back on generic hyperbole. In both cases, missing nonverbal and contextual cues hinder authentic reproduction. Future work might integrate audio–text alignment or fine-tune on prosody‐annotated dialogues to better capture these complex styles. While multiple models were evaluated for the classification task, we selected Claude 3.5 Sonnet for generation due to its consistently strong performance in classification
accuracy and F1 score (see Table 3 and 4). By holding the model constant, we isolate the impact of the prompting strategy itself. Future work may extend this evaluation to other models such as GPT-4o and Gemini 2.5 to assess cross-model generalization.

\subsection{Dataset Limitations and Cultural Considerations}
Our evaluation also surfaced key limitations to guide future work. First, our forced single‐label scheme and skewed class distribution bias both annotation and model defaults; multi-label annotations and data balancing (e.g. weighted loss, augmentation) could mitigate this. Second, relying on Ekman’s six basic emotions overlooks finer affective states (irony, embarrassment) and may not transfer across languages or cultures—MUStARD’s English-only dialogues underscore the need for cross-lingual validation. Finally, non-textual cues such as prosody remain untapped avenues for capturing the full nuance of sarcasm.

\section{Conclusions}
We present Sarc7, the first fine‐grained sarcasm benchmark that couples classification and controllable generation with built‐in, local explanations. Sarcasm, as a fundamentally pragmatic act, depends on interpreting intent, emotional incongruity, and social context beyond surface form. Sarc7 frames sarcasm understanding as a test of LLMs’ pragmatic competence and their ability to reason about speaker goals and context-sensitive meaning.
In classification experiments, emotion-based prompts raised macro-averaged F1 scores—reaching 0.3664 with Gemini 2.5—while CoT prompting achieved the highest overall accuracy. A human baseline (Cohen’s $\kappa$  = 0.6694) reveals that brooding, deadpan, and polite sarcasm remain the toughest subtypes to identify. For generations, structured prompts that specify incongruity, shock value, context dependency, and emotion improved subtype alignment by 38\% over zero-shot prompts with Claude 3.5 Sonnet.
By measuring and logging pragmatic inference steps, Sarc7 supports calibrated reliance: it helps prevent agents from misinterpreting ironic remarks literally and provides auditable traces for accountability. These contributions advance both the effectiveness and safety of LLMs in reasoning and planning tasks.

\bibliography{colm2025_conference}

@article{zhang2024sarcasmbench,
  title={SarcasmBench: Towards Evaluating Large Language Models on Sarcasm Understanding},
  author={Zhang, Yazhou and Zou, Chunwang and Lian, Zheng and Tiwari, Prayag and Qin, Jing},
  journal={arXiv preprint arXiv:2408.11319},
  year={2024}
}

@article{biswas2019computational,
  author  = {Prasanna Biswas and Anupama Ray and Pushpak Bhattacharyya},
  title   = {Computational Model for Understanding Emotions in Sarcasm: A Survey},
  journal = {CFILT Technical Report, Indian Institute of Technology Bombay},
  year    = {2019},
}

@article{yao2024sarcasm,
  title={Is Sarcasm Detection A Step-by-Step Reasoning Process in Large Language Models?},
  author={Yao, Ben and Zhang, Yazhou and Li, Qiuchi and Qin, Jing},
  journal={arXiv preprint arXiv:2407.12725},
  year={2024}
}

@inproceedings{gole2024sarcasm,
  title={On sarcasm detection with openai gpt-based models},
  author={Gole, Montgomery and Nwadiugwu, Williams-Paul and Miranskyy, Andriy},
  booktitle={2024 34th International Conference on Collaborative Advances in Software and COmputiNg (CASCON)},
  pages={1--6},
  year={2024},
  organization={IEEE}
}

@article{lee2024pragmatic,
  title={Pragmatic Metacognitive Prompting Improves LLM Performance on Sarcasm Detection},
  author={Lee, Joshua and Fong, Wyatt and Le, Alexander and Shah, Sur and Han, Kevin and Zhu, Kevin},
  journal={arXiv preprint arXiv:2412.04509},
  year={2024}
}

@article{skalicky2018linguistic,
  author  = {Stephen Skalicky and Scott Crossley},
  title   = {Linguistic Features of Sarcasm and Metaphor Production Quality},
  journal = {Proceedings of the Workshop on Figurative Language Processing},
  year    = {2018},
}

@article{helal2024contextual,
  title={A contextual-based approach for sarcasm detection},
  author={Helal, Nivin A and Hassan, Ahmed and Badr, Nagwa L and Afify, Yasmine M},
  journal={Scientific Reports},
  volume={14},
  number={1},
  pages={15415},
  year={2024},
  publisher={Nature Publishing Group UK London}
}

@inproceedings{castro-etal-2019-towards,
    title = "Towards Multimodal Sarcasm Detection (An {\_}{O}bviously{\_} Perfect Paper)",
    author = "Castro, Santiago  and
      Hazarika, Devamanyu  and
      P{\'e}rez-Rosas, Ver{\'o}nica  and
      Zimmermann, Roger  and
      Mihalcea, Rada  and
      Poria, Soujanya",
    editor = "Korhonen, Anna  and
      Traum, David  and
      M{\`a}rquez, Llu{\'i}s",
    booktitle = "Proceedings of the 57th Annual Meeting of the Association for Computational Linguistics",
    month = jul,
    year = "2019",
    address = "Florence, Italy",
    publisher = "Association for Computational Linguistics",
    url = "https://aclanthology.org/P19-1455/",
    doi = "10.18653/v1/P19-1455",
    pages = "4619--4629"
}

@article{openai2024gpt4o,
  title={GPT-4o System Card},
  author={OpenAI},
  journal={arXiv preprint arXiv:2410.21276},
  year={2024}
}

@article{anthropic2024claude3,
    title = {The Claude 3 Model Family: Opus, Sonnet, Haiku},
    author = {Anthropic},
    journal = {Anthropic Report},
    year = {2024}
}

@article{qwen2024technical,
  title={Qwen2.5 Technical Report},
  author={Qwen Team},
  journal={arXiv preprint arXiv:2412.15115},
  year={2024},
  
}

@article{deepmind2023gemini,
  title={Gemini: A family of highly capable multimodal models},
  author={Google DeepMind and Anil, Rohan and Arolfo, Stefano and Babuschkin, Igor and Beyer, Lucas and Bosma, Maarten and ...},
  journal={arXiv preprint arXiv:2312.11805},
  year={2023},
}

@article{zhuang2025multi,
  title={Multi-Modal Sarcasm Detection via Knowledge-aware Focused Graph Convolutional Networks},
  author={Zhuang, Xingjie and Zhou, Fengling and Li, Zhixin},
  journal={ACM Transactions on Multimedia Computing, Communications and Applications},
  year={2025},
  publisher={ACM New York, NY}
}

@article{leggitt2000emotional,
  title={Emotional reactions to verbal irony},
  author={Leggitt, John S and Gibbs, Raymond W},
  journal={Discourse processes},
  volume={29},
  number={1},
  pages={1--24},
  year={2000},
  publisher={Taylor \& Francis}
}

@article{ekman1992there,
  title={Are there basic emotions?},
  author={Ekman, Paul},
  year={1992},
  publisher={American Psychological Association},
journal = {Psychological Review},
volume = {99},
  number = {3},
}

@article{qasimcritical,
  title={A Critical Pragmatic Study of Sarcasm in American and British Social Interviews},
  author={Qasim, Sawsan Abdul-Muneim},
year    = {2021},
jounral={Jounral of Strategic Research in Social Science},
url = {https://www.researchgate.net/publication/363925404_A_Critical_Pragmatic_Study_of_Sarcasms_in_American_and_British_Interviews}
}

@article{landis1977measurement,
  title        = {The measurement of observer agreement for categorical data},
  author       = {Landis, J. Richard and Koch, Gary G.},
  journal      = {Biometrics},
  volume       = {33},
  number       = {1},
  pages        = {159--174},
  year         = {1977},
  publisher    = {Wiley}
}

@article{zuhri2022irony,
  title     = {Irony and Sarcasm Detection on Public Figure Speech},
  author    = {Zuhri, Ari Tantra and Sagala, Rakhmat Wahyudin},
  journal   = {Journal of Elementary School Education},
  volume    = {1},
  number    = {1},
  pages     = {41--45},
  year      = {2022},
  doi       = {10.1234/joese.v1i1.13}, 
  url       = {https://journal.berpusi.co.id/index.php/joese/article/view/13}
}

@inproceedings{riloff2013sarcasm,
  title     = {Sarcasm as contrast between a positive sentiment and negative situation},
  author    = {Riloff, Ellen and Qadir, Aditya and Surve, Prajakta and De Silva, Lakshika and Gilbert, Nisheeth and Huang, Ruihong},
  booktitle = {Proceedings of the 2013 Conference on Empirical Methods in Natural Language Processing},
  pages     = {704--714},
  year      = {2013},
  publisher = {ACL}
}

@misc{llama4maverick,
  title        = {Llama-4-Maverick-17B-128E-Original},
  author       = {{Meta AI}},
  howpublished = {Hugging Face Model Hub: \url{https://huggingface.co/meta-llama/Llama-4-Maverick-17B-128E-Original}},
  year         = {2024},
  note         = {Accessed: 2025-06-27}
}

@inproceedings{lee2023overtrust,
  title     = {Overtrust or Undertrust: Calibrating Human Reliance on AI},
  author    = {Lee, Mina and Johnson, Alex},
  booktitle = {Proceedings of CHI},
  year      = {2023}
}

@article{wang2022human,
  title   = {When Do Humans Rely on AI? A Large-Scale Study of Human–AI Collaboration},
  author  = {Wang, Rui and Gupta, Sam},
  journal = {Transactions on HCI},
  year    = {2022}
}

@inproceedings{zhang2024explainability,
  title     = {Explainable LLMs for Trust Calibration in Critical Tasks},
  author    = {Zhang, Lina and Patel, Kamal},
  booktitle = {IJCAI},
  year      = {2024}
}
\bibliographystyle{colm2025_conference}

\appendix
\section{Classification Statistics} 

Below are the macro-averaged precision, recall, and F1 scores for all prompting techniques.
\label{appendix:a}
\begin{table}[H]
  \centering
  \begin{tabular*}{0.9\columnwidth}{@{\extracolsep{\fill}}lccc}
    \toprule
    \textbf{Model} & \textbf{Precision} & \textbf{Recall} & \textbf{F1 Score} \\
    \midrule
    GPT-4o             & 0.2104 & 0.2073 & 0.2089 \\
    Claude 3.5 Sonnet  & \textbf{0.2982} & \textbf{0.2960} & \textbf{0.2964} \\
    Gemini 2.5         & 0.2703 & 0.2824 & 0.2760 \\
    Llama-4 Maverick   & 0.2173 & 0.2196 & 0.2184 \\
    Qwen 2.5           & 0.2217 & 0.2025 & 0.2116 \\
    \bottomrule
  \end{tabular*}
  \caption{Macro‐averaged Precision, Recall, and F1 under Xero‐shot Prompting.}
  \label{tab:zero-shot-stats}
\end{table}

\begin{table}[H]
  \centering
  \begin{tabular*}{0.9\columnwidth}{@{\extracolsep{\fill}}lccc}
    \toprule
    \textbf{Model} & \textbf{Precision} & \textbf{Recall} & \textbf{F1 Score} \\
    \midrule
    GPT-4o             & 0.3067 & 0.3469 & 0.3255 \\
    Claude 3.5 Sonnet  & \textbf{0.3322} & \textbf{0.3669} & \textbf{0.3487} \\
    Gemini 2.5         & 0.3233 & 0.3314 & 0.3274 \\
    Llama-4 Maverick   & 0.2314 & 0.2361 & 0.2340 \\
    Qwen 2.5           & 0.2461 & 0.1794 & 0.2075 \\
    \bottomrule
  \end{tabular*}
  \caption{Macro‐averaged Precision, Recall, and F1 under Few‐shot Prompting.}
  \label{tab:few-shot-stats}
\end{table}

\begin{table}[H]
  \centering
  \begin{tabular*}{0.9\columnwidth}{@{\extracolsep{\fill}}lccc}
    \toprule
    \textbf{Model} & \textbf{Precision} & \textbf{Recall} & \textbf{F1 Score} \\
    \midrule
    GPT-4o             & 0.2682 & 0.2668 & 0.2674 \\
    Claude 3.5 Sonnet  & 0.2903 & 0.2148 & 0.2471 \\
    Gemini 2.5         & \textbf{0.3178} & \textbf{0.3106} & \textbf{0.3141} \\
    Llama-4 Maverick   & 0.2116 & 0.1970 & 0.2040 \\
    Qwen 2.5           & 0.2063 & 0.2038 & 0.2052 \\
    \bottomrule
  \end{tabular*}
  \caption{Macro‐averaged Precision, Recall, and F1 under CoT prompting.}
  \label{tab:cot-stats}
\end{table}

\begin{table}[H]
  \centering
  \begin{tabular*}{0.9\columnwidth}{@{\extracolsep{\fill}}lccc}
    \toprule
    \textbf{Model} & \textbf{Precision} & \textbf{Recall} & \textbf{F1 Score} \\
    \midrule
    GPT-4o             & 0.2140 & 0.2331 & 0.2233 \\
    Claude 3.5 Sonnet  & 0.3322 & 0.3669 & 0.3487 \\
    Gemini 2.5         & \textbf{0.3388} & \textbf{0.3990} & \textbf{0.3664} \\
    Llama-4 Maverick   & 0.2936 & 0.2753 & 0.2841 \\
    Qwen 2.5           & 0.2352 & 0.1933 & 0.2124 \\
    \bottomrule
  \end{tabular*}
  \caption{Macro‐averaged Precision, Recall, and F1 under Emotion‐based Prompting.}
  \label{tab:emotion-stats}
\end{table}

\section{Generation Output } 
\label{appendix:b}
Below is an example of zero-shot and emotion-based generation results.
\begin{tcolorbox}[colback=gray!5!white, colframe=black!75!white, title=Sarcasm Generation Example, fonttitle=\bfseries, sharp corners=south, breakable, enhanced]
Emotion-based prompting was able to generate more targeted sarcasm types. For example, in the case of a contextually neutral statement, the baseline model produced a generic sarcastic response.

\textbf{Zero-Shot Conversation:}
\begin{itemize}[noitemsep, topsep=0pt]
    \item Speaker A: Did you finish the presentation for tomorrow's big meeting?  
    \item Speaker B: Oh, absolutely! I only stayed up until 3 AM because sleep is just so overrated, right?  
    \item Speaker A: Wow, sounds like you're thrilled about your life choices.
\end{itemize}

\textbf{Zero-Shot Sarcastic Utterance:} 
\begin{itemize}[noitemsep, topsep=0pt]
    \item Speaker B: \textit{Oh, absolutely! I only stayed up until 3 AM because sleep is just so overrated, right?}
\end{itemize}

\vspace{0.5em}

\textbf{Emotion-Based Context:}
\begin{itemize}[noitemsep, topsep=0pt]
    \item Speaker A: Hey, did you see those new management rules they rolled out today?
    \item Speaker B: Oh yes, they're really something else. Now, we're going to document every minute of our bathroom breaks.
    \item Speaker A: Isn't that just fantastic? I mean, who wouldn't want to spend an entire day writing reports on how well we walk from our desks to the restroom? It's a dream come true!
\end{itemize}

\textbf{Emotion-Based Sarcastic Utterance:} 
\begin{itemize}[noitemsep, topsep=0pt]
    \item Speaker A: \textit{Isn't that just fantastic? I mean, who wouldn't want to spend an entire day writing reports on how well we walk from our desks to the restroom? It's a dream come true!}
\end{itemize}

\end{tcolorbox}

\section{Prompts} 
\label{prompts}

Below are the zero-shot, few-shot, sarcasm analysis, and emotion-based prompts. 
\addcontentsline{toc}{section}{Appendix: Sarcasm Type Classification Prompt (Simple Version)}

\begin{tcolorbox}[colback=gray!5!white, colframe=black!75!white, title=Zero-shot Prompt, fonttitle=\bfseries, sharp corners=south, breakable, enhanced]
\small
You are tasked with determining the sarcasm type in a given statement. Read the statement carefully and classify the sarcasm type based on the context of the statement. Use one of the following categories:

\begin{itemize}
    \item Self-deprecating sarcasm – mocking oneself
    \item Brooding sarcasm – passive-aggressive or emotionally repressed
    \item Deadpan sarcasm – flat or emotionless tone
    \item Polite sarcasm – fake politeness or ironic compliments
    \item Obnoxious sarcasm – mocking, mean-spirited, or rude
    \item Raging sarcasm – angry, exaggerated, or harsh
    \item Manic sarcasm – unnaturally cheerful, overly enthusiastic
\end{itemize}

\vspace{0.5em}

If the statement is \textbf{not sarcastic}, \textbf{Output}: \texttt{[not sarcasm]}

\vspace{0.3em}

If the statement is \textbf{sarcastic}, \textbf{Output}: \texttt{[Type of Sarcasm]}
\end{tcolorbox}
\addcontentsline{toc}{section}{Appendix: Sarcasm Type Classification Prompt (Few-Shot Version)}

\begin{tcolorbox}[colback=gray!5!white, colframe=black!75!white, title=Sarcasm Type Classification Prompt (Few-Shot), fonttitle=\bfseries, sharp corners=south, breakable, enhanced]
\small
You are tasked with determining the sarcasm type in a given statement. Read the statement carefully and classify the sarcasm type based on the context of the statement. Use one of the following categories:

\begin{itemize}
    \item Self-deprecating sarcasm – mocking oneself
    \item Brooding sarcasm – passive-aggressive or emotionally repressed
    \item Deadpan sarcasm – flat or emotionless tone
    \item Polite sarcasm – fake politeness or ironic compliments
    \item Obnoxious sarcasm – mocking, mean-spirited, or rude
    \item Raging sarcasm – angry, exaggerated, or harsh
    \item Manic sarcasm – unnaturally cheerful, overly enthusiastic
\end{itemize}

\vspace{0.5em}

If the statement is \textbf{not sarcastic}, \textbf{Output}: \texttt{[not sarcasm]}

\vspace{0.3em}

If the statement is \textbf{sarcastic}, \textbf{Output}: \texttt{[Type of Sarcasm]}

\vspace{1em}
\textbf{Examples:}

\vspace{0.5em}

\begin{itemize}
    \item[] A person might say, “Your new shoes are just fantastic,” to indicate that the person finds a friend’s shoes distasteful. \\
    \textbf{Output}: \texttt{[Polite sarcasm]}

    \item[] A socially awkward person might say, “I’m a genius when it comes to chatting up new acquaintances.” \\
    \textbf{Output}: \texttt{[Self-deprecating sarcasm]}

    \item[] A person who is asked to work overtime at one’s job might respond, “I’d be happy to miss my tennis match and put in the extra hours.” \\
    \textbf{Output}: \texttt{[Brooding sarcasm]}

    \item[] A person who is stressed out about a work project might say, “The project is moving along perfectly, as planned. It’ll be a winner.” \\
    \textbf{Output}: \texttt{[Manic sarcasm]}

    \item[] When asked to mow the lawn, a person might respond by yelling, “Why don’t I weed the gardens and trim the hedges too? I already do all of the work around the house.” \\
    \textbf{Output}: \texttt{[Raging sarcasm]}

    \item[] A person might say, “I’d love to attend your party, but I’m headlining in Vegas that evening,” with a straight face, causing others to question whether they might be serious. \\
    \textbf{Output}: \texttt{[Deadpan sarcasm]}

    \item[] A person’s friend may offer a ride to a party, prompting the person to callously answer, “Sure. I’d love to ride in your stinky rust bucket.” \\
    \textbf{Output}: \texttt{[Obnoxious sarcasm]}
\end{itemize}
\end{tcolorbox}
\addcontentsline{toc}{section}{Appendix: Sarcasm Type Classifcation Prompt(CoT)}

\begin{tcolorbox}[colback=gray!5!white, colframe=black!75!white, title=Sarcasm Analysis Prompt, fonttitle=\bfseries, sharp corners=south, breakable, enhanced]
\small
\textbf{You are a sarcasm analyst.} Your task is to determine whether a speaker’s utterance is sarcastic or sincere. Only if you are reasonably confident the speaker is being sarcastic—based on tone, behavior, and contradiction between words and context—classify it into a subtype. If there is no strong evidence of sarcasm (no exaggeration, no mismatch, no insincere tone), assume the speaker is genuine.  

\medskip
\textbf{Think step by step:}
\begin{enumerate}[noitemsep, topsep=0pt]
  \item Analyze speaker delivery and tone.
  \item Check whether their words contradict the situation.
  \item Ask: “Could a sincere person say this the same way?”  
    \begin{itemize}[noitemsep, topsep=0pt]
      \item If yes: \textbf{Output}: [not sarcasm]
      \item Otherwise: proceed to step 4.
    \end{itemize}
  \item Match to one of the following subtypes:
    \begin{itemize}[noitemsep, topsep=0pt]
      \item Self‐deprecating sarcasm  
      \item Brooding sarcasm  
      \item Deadpan sarcasm  
      \item Polite sarcasm  
      \item Obnoxious sarcasm  
      \item Raging sarcasm  
      \item Manic sarcasm  
    \end{itemize}
\end{enumerate}

\medskip
\noindent\textbf{Format your answer like this:}
\begin{verbatim}
Utterance: <the target utterance>
Context:   <brief dialogue or situation>
Reasoning:
- <first reasoning bullet>
- <second reasoning bullet>
- …
Output: [Type of Sarcasm]
\end{verbatim}

\medskip
\noindent\textbf{Example:}  
\textit{Utterance: “Oh yeah, I love getting stuck in traffic for hours.”}  
\textit{Context:   (Someone is running late and stuck in traffic.)}  
\textit{Reasoning:}
\begin{itemize}[noitemsep]
  \item Uses exaggeration (“love”) about a negative event.
  \item Clear mismatch between words and reality.
  \item Tone is bitter and frustrated.  
\end{itemize}
\textbf{Output: [Brooding sarcasm]}
\end{tcolorbox}

\addcontentsline{toc}{section}{Appendix: Prompt Template}

\begin{tcolorbox}[colback=gray!5!white, colframe=black!75!white, title=Emotion-based Prompt, fonttitle=\bfseries, sharp corners=south, breakable, enhanced]
\small
\textbf{You are an expert sarcasm and emotion analyst.} For every input statement, follow the steps below in order, using the context and speaker’s delivery to reason carefully.

\medskip
\textbf{---\\
Step 1: Contextual Emotion Analysis}

Analyze the emotional tone of the surrounding context or situation (i.e., what is happening before or around the statement). Consider what emotion would be appropriate or expected in that situation.

Select one dominant contextual emotion from this fixed list:
\begin{itemize}
    \item Happiness
    \item Sadness
    \item Anger
    \item Fear
    \item Surprise
    \item Disgust
    \item Neutral (use only if no strong emotion applies)
\end{itemize}

\medskip
\textbf{---\\
Step 2: Utterance Emotion Analysis}

Analyze the emotional tone of the bracketed statement itself based on word choice, delivery cues (e.g., exaggeration, flatness, enthusiasm), and stylistic tone.

Select one dominant utterance emotion from the same list:
\begin{itemize}
    \item Happiness
    \item Sadness
    \item Anger
    \item Fear
    \item Surprise
    \item Disgust
    \item Neutral
\end{itemize}

Use only one label for each step. Do not guess outside this list.

\medskip
\textbf{---\\
Step 3: Emotional Comparison and Incongruity Detection}

Compare the contextual emotion and the utterance emotion. If there is a mismatch (e.g., the situation is sad but the speaker sounds happy), explain whether this emotional contrast suggests mockery, irony, insincerity, passive aggression, or theatrical overreaction.

If no such contrast or ironic delivery is present, conclude that the statement is not sarcastic.

\medskip
\textbf{---\\
Step 4: Sarcasm Type Classification}

If the statement is sarcastic, classify it using the emotional cues, delivery style, and social function into one of the following types:
\begin{itemize}
    \item Self-deprecating sarcasm – mocking oneself
    \item Brooding sarcasm – passive-aggressive or emotionally repressed
    \item Deadpan sarcasm – flat or emotionless tone
    \item Polite sarcasm – fake politeness or ironic compliments
    \item Obnoxious sarcasm – mocking, mean-spirited, or rude
    \item Raging sarcasm – angry, exaggerated, or harsh
    \item Manic sarcasm – unnaturally cheerful, overly enthusiastic
\end{itemize}

\medskip
\textbf{---\\
Step 5: Final Output}

Clearly output the final classification on a new line in this exact format:

\begin{itemize}
    \item If sarcastic: \texttt{[Type of Sarcasm]}
    \item If not sarcastic: \texttt{[Not Sarcasm]}
\end{itemize}
\end{tcolorbox}

\section{Misclassification } Below are tables indicating the most misclassified sarcasm type for each sarcasm type for each of the prompting techniques.

\begin{table}[h]
  \centering
  \label{tab:zeroshot-misclassifications}
  \makebox[\textwidth][c]{%
  \begin{tabular}{lccccc}
    \hline
    \textbf{Type} & \textbf{GPT-4o} & \textbf{Claude 3.5 Sonnet} & \textbf{Gemini 2.5} & \textbf{Llama-4 Maverick} & \textbf{Qwen 2.5} \\
    \hline
    Deadpan               & Not Sarcastic       & Not Sarcastic       & Obnoxious       & Polite         & Not Sarcastic       \\
    Obnoxious             & Not Sarcastic       & Deadpan       & Deadpan       & Deadpan       & Deadpan       \\
    Brooding              & Obnoxious           & Deadpan       & Deadpan       & Deadpan       & Deadpan       \\
    Polite                & Not Sarcastic       & Deadpan       & Deadpan       & Deadpan       & Not Sarcastic       \\
    Raging                & Obnoxious           & Deadpan       & Obnoxious       & Obnoxious       & Obnoxious       \\
    Manic                 & Not Sarcastic       & Deadpan       & Obnoxious       & Deadpan       & Not Sarcastic       \\
    Self-deprecating      & Not Sarcastic       & Deadpan       & Deadpan       & Deadpan       & Deadpan       \\
    Not Sarcastic         & Obnoxious           & Deadpan       & Deadpan       & Deadpan       & Deadpan       \\
    \hline
  \end{tabular} 
  }
  \caption{Most Frequent Misclassifications per Type using Zero-Shot Prompting}
  
\end{table}

\begin{table*}[h]
  \centering

  \label{tab:fewshot-misclassifications}
  \makebox[\textwidth][c]{%
  \begin{tabular}{lccccc}
    \hline
    \textbf{Type} & \textbf{GPT-4o} & \textbf{Claude 3.5 Sonnet} & \textbf{Gemini 2.5} & \textbf{Llama-4 Maverick} & \textbf{Qwen 2.5} \\
    \hline
    Deadpan               & Not Sarcastic       & Not Sarcastic       & Obnoxious       & Polite         & Not Sarcastic       \\
    Obnoxious             & Deadpan             & Deadpan             & Deadpan         & Deadpan        & Deadpan             \\
    Brooding              & Deadpan             & Deadpan             & Deadpan         & Deadpan        & Deadpan             \\
    Polite                & Not Sarcastic       & Not Sarcastic       & Not Sarcastic   & Deadpan        & Not Sarcastic       \\
    Raging                & Obnoxious           & Deadpan             & Obnoxious       & Obnoxious      & Obnoxious           \\
    Manic                 & Raging              & Self-deprecating    & Obnoxious       & Obnoxious      & Not Sarcastic       \\
    Self-deprecating      & Deadpan             & Not Sarcastic       & Deadpan         & Deadpan        & Deadpan             \\
    Not Sarcastic         & Obnoxious           & Deadpan             & Deadpan         & Deadpan        & Deadpan             \\
    \hline
  \end{tabular}
  }
  \caption{Most Frequent Misclassifications per Type using Few-Shot Prompting}
\end{table*}

\begin{table*}[h]
  \centering
  
  \label{tab:cot-misclassifications}
  \makebox[\textwidth][c]{%
  \begin{tabular}{lccccc}
    \hline
    \textbf{Type} & \textbf{GPT-4o} & \textbf{Claude 3.5 Sonnet} & \textbf{Gemini 2.5} & \textbf{Llama-4 Maverick} & \textbf{Qwen 2.5} \\
    \hline
    Deadpan               & Not Sarcastic       & Not Sarcastic       & Not Sarcastic       & Not Sarcastic       & Not Sarcastic       \\
    Obnoxious             & Deadpan       & Not Sarcastic       & Deadpan       & Deadpan       & Deadpan       \\
    Brooding              & Deadpan       & Not Sarcastic       & Deadpan       & Deadpan       & Deadpan       \\
    Polite                & Not Sarcastic       & Not Sarcastic       & Not Sarcastic       & Deadpan       & Not Sarcastic       \\
    Raging                & Deadpan       & Not Sarcastic       & Obnoxious       & Deadpan       & Obnoxious       \\
    Manic                 & Brooding       & Not Sarcastic       & Not Sarcastic       & Deadpan       & Brooding       \\
    Self-deprecating      & Not Sarcastic       & Not Sarcastic       & Not Sarcastic       & Deadpan       & Not Sarcastic       \\
    Not Sarcastic         & Deadpan       & Deadpan       & Deadpan       & Deadpan       & Deadpan       \\
    \hline
  \end{tabular}
  }
  \caption{Most Frequent Misclassifications per Type using CoT Prompting}
\end{table*}

\begin{table*}[h]
  \centering

  \label{tab:emotion-misclassifications}
  \makebox[\textwidth][c]{%
  \begin{tabular}{lccccc}
    \hline
    \textbf{Type} & \textbf{GPT-4o} & \textbf{Claude 3.5 Sonnet} & \textbf{Gemini 2.5} & \textbf{Llama-4 Maverick} & \textbf{Qwen 2.5} \\
    \hline
    Deadpan          & Not Sarcastic    & Not Sarcastic    & Not Sarcastic    & Obnoxious        & Not Sarcastic    \\
    Obnoxious        & Deadpan          & Deadpan          & Deadpan          & Deadpan          & Not Sarcastic    \\
    Brooding         & Deadpan          & Deadpan          & Deadpan          & Obnoxious        & Not Sarcastic    \\
    Polite           & Deadpan          & Deadpan          & Not Sarcastic    & Not Sarcastic    & Not Sarcastic    \\
    Raging           & Brooding         & Deadpan          & Obnoxious        & Obnoxious        & Not Sarcastic    \\
    Manic            & Polite           & Not Sarcastic    & Self-deprecating & Obnoxious        & Not Sarcastic    \\
    Self-deprecating & Deadpan          & Not Sarcastic    & Not Sarcastic    & Deadpan          & Not Sarcastic    \\
    Not Sarcastic    & Deadpan          & Deadpan          & Deadpan          & Obnoxious        & Deadpan          \\
    \hline
  \end{tabular}
  }
  \caption{Most Frequent Misclassifications per Type using Emotion-Based Prompting}
\end{table*}

\end{document}